\documentclass[preprint,12pt,3p]{elsarticle}

\usepackage{amssymb}
\usepackage[utf8]{inputenc}
\usepackage{graphicx}
\usepackage{subfig}
\usepackage{multirow}
\usepackage{verbatim}
\usepackage{enumitem}
\usepackage{url}
\usepackage{etoolbox}
\usepackage{booktabs}
\usepackage{tocbibind}
\usepackage[toc,page]{appendix}
\makeatletter
\patchcmd{\Ginclude@eps}{"#1"}{#1}{}{}
\newcommand*{\rom}[1]{\expandafter\@slowromancap\romannumeral #1@}
\makeatother

\journal{Neural Networks}

\begin{document}

\begin{frontmatter}

\title{Learning in the Machine: To Share or Not to Share?}

\author[label1,label2]{Jordan Ott} 
\ead{jott1@uci.edu}
\address[label1]{Fowler School of Engineering\\ 
Chapman University}
\address[label2]{Department of Computer Science\\
Bren School of Information and Computer Science\\
University of California, Irvine}

\author[label1]{Erik Linstead\corref{cor1}}
\ead{linstead@chapman.edu}
\cortext[cor1]{Corresponding author}

\author[label1]{Nicholas LaHaye}
\ead{lahay100@mail.chapman.edu }

\author[label2]{Pierre Baldi}
\ead{pfbaldi@ics.uci.edu}

\begin{abstract}
Weight-sharing is one of the pillars behind Convolutional Neural Networks and their successes. However, in physical neural systems such as the brain, weight-sharing is implausible. This discrepancy raises the fundamental question of whether weight-sharing is necessary. If so, to which degree of precision? If not, what are the alternatives? The goal of this study is to investigate these questions, primarily through simulations where the weight-sharing assumption is relaxed. Taking inspiration from neural circuitry, we explore the use of Free Convolutional Networks and neurons with variable connection patterns. Using Free Convolutional Networks, we show that while weight-sharing is a pragmatic optimization approach, it is not a necessity in computer vision applications. Furthermore, Free Convolutional Networks match the performance observed in standard architectures when trained using properly translated data (akin to video). Under the assumption of translationally augmented data, Free Convolutional Networks learn translationally invariant representations that yield an approximate form of weight-sharing.
\end{abstract}


\end{frontmatter}



\section{Introduction}
Digital simulations of neural networks are successful in many applications but rely on a fantasy where neurons and synaptic weights are objects stored in digital computer memories. This fantasy often obfuscates some fundamental principles of computing in native neural systems. To remedy this obfuscation, learning in the machine refers to a general approach for studying neural computations. In this approach, the physical constraints of physical neural systems, such as brains or neuromorphic chips, are taken into consideration. When applied to single neurons, learning in the machine can lead, for instance, to the discovery of dropout \cite{srivastava2014dropout,baldidropout14}. When applied to synapses, learning in the machine can lead, for instance, to the discovery of local learning \cite{baldi2016local} and random backpropagation \cite{lillicrap2016random,baldiRBP2016AI,baldiSymmetries2017}. Moreover, when applied to layers of neurons, as we do in this paper, learning in the machine leads one to question the fundamental assumption of weight-sharing behind convolutional neural networks (CNNs). 
  
The technique of weight-sharing, whereby different synaptic connections share the same strength, is a widely used and successful technique in neural networks and deep learning. This paradigm is particularly true in computer vision where weight-sharing is one of the pillars behind convolutional neural networks and their successes. In any physical neural system, for instance, carbon- or silicon-based, exact sharing of connections strengths over spatial distances is difficult to realize, especially on a massive 3D scale. In physical systems, not only is it difficult to create identical weights at a given time point, but it is also challenging to maintain the identity over time. During phases of development and learning the weights may be changing rapidly. During more mature stages weights must retain their integrity against the microscopic, entropic forces surrounding any physical synapse. Furthermore, given the exquisitely complex geometry of neuronal dendritic trees and axon arborizations, it is implausible to form large arrays of neurons with identically translated connection patterns. In short, not only is it challenging to share weights exactly, but it is also difficult to exactly share the same connection patterns.

While weight-sharing has proven to be very useful in computer vision and other applications, it is extremely implausible in biological and other physical systems. This discrepancy raises the fundamental question of whether weight-sharing is a strict prerequisite for convolution-based deep learning, or if similar levels of learning are possible without it. In particular, we consider the following research questions:

\begin{enumerate}
    \item Is weight-sharing necessary to prevent overfitting?
    \item Is weight-sharing necessary to ensure translational invariant recognition?
    \item Can acceptable classification performance be achieved without weight-sharing?
    \item Does approximate or exact weight-sharing emerge in a natural way?
\end{enumerate}

The goal of this study is to investigate these research questions, primarily through simulations where the weight-sharing assumption is relaxed. In total, the answer to these questions provide new insight into whether weight-sharing is vital for convolutional architectures.

\section{Origins and Functions of Weight-Sharing}
Before addressing the question of its necessity, it is useful to review the origins and functions of weight-sharing. Hubel and Wiesel's neurophysiological work \cite{hubel1962receptive} on the cat visual cortex was the inception of weight-sharing. These experiments suggested the existence of entire arrays of neurons dedicated to implementing simple operations, such as edge detection and other Gabor filters, at all possible image locations. Fukushima systematically used the ideas proposed by Hubel and Wiesel to create the neocognitron \cite{fukushima1980neocognitron} in computer vision. Primarily a convolutional neural network architecture with Hebbian learning. However, Hebbian Learning alone applied to a feedforward CNN cannot solve vision tasks \cite{baldi2016local}. Solving vision tasks requires feedback channels and learning algorithms for transmitting target information to the deep synapses. A job that is precisely achieved by backpropagation, or stochastic gradient descent. Successful CNNs for vision problems, trained via backpropagation, were developed in the late 80s and 90s  \cite{lecun90,baldi93_fingerprint,schmidhuber2015deep}. 

Substantial improvements in the size of the training sets and available computing power have led to a new wave of successful implementations in recent years, \cite{krizhevsky2012imagenet,Szegedy15_GoogLenet,srivastava2015training,He15_resNet}, as well as applications to a variety of specific domains, ranging from biomedical images \cite{Ciresan12_membraneSegmentation,gulshan2016development,wang2017multi,wang2017detecting,esteva2017dermatologist,baldicolonoscopy2017} to particle physics \cite{baldi2016jet,Aurisano_2016jvx,2399-6528-1-2-025001} and video analysis \cite{ott2018deep, tompson2014real, ott2018learning, tran2015learning}.
Older \cite{zipser1988back} as well as more recent work \cite{Cadieu_2014_PLOS,Yamins10062014} has also shown that not only do convolutional neural networks rival the object category recognition accuracy of the primate cortex but also seem to provide the best match to biological neural responses, at least at some coarse level of analysis. 

It is worth pointing out that weight-sharing is sometimes used in other settings, for instance when Siamese Networks are used to process and compare objects \cite{bromley-93,baldi93_fingerprint,kayala2012reactionpredictor}, which also includes Siamese CNNs for images. Finally, a different kind of weight-sharing that will not concern us here, when a recurrent network is unfolded in time. In this case, weight-sharing occurs over time and not over space \cite{hochreiter1997long, cho2014learning}.

Biologically motivated architectures have become an import focal point of deep learning as of late \cite{bartunov2018assessing, baldi2018learning, baldi2017learning,samadi2017deep, lillicrap2016random, baldi2018recirc}. Most recently \cite{bartunov2018assessing} investigated how biologically motivated deep learning algorithms scale to more massive datasets. Locally connected networks were included in their simulations, where each weight kernel interacts with \textit{local} features, not the entire input.  However, the focus of the paper is on how learning algorithms such as feedback alignment \cite{lillicrap2016random} and target propagation \cite{lee2015difference} scale to more significant image recognition problems. Not addressed in the paper is the problem of learning translationally invariant representations, variable connection patterns, and the emergence of approximate weight-sharing. Nor was there in-depth analysis comparing networks with weight-sharing to those without. 

Two primary but different purposes are typically associated with weight-sharing. The first is to reduce the number of free parameters that need to be stored or updated during learning. This requirement can be important in applications where storage space is limited (i.e., cell phones), or where training data is limited, and overfitting is a danger. It is important to remember, however, that in convolutional architectures the local connectivity of neurons contributes at least as much to parameter reduction as weight-sharing. In other words, weight-sharing is not the only way to shrink the parameter space. The second function of weight-sharing is to apply the same operation at different locations of the input data, to process the data uniformly and provide a basis for invariance, typically translation invariant recognition in CNN architectures.

\section{Free Convolutional Networks}

\begin{figure}
\center
  \includegraphics[width=40mm]{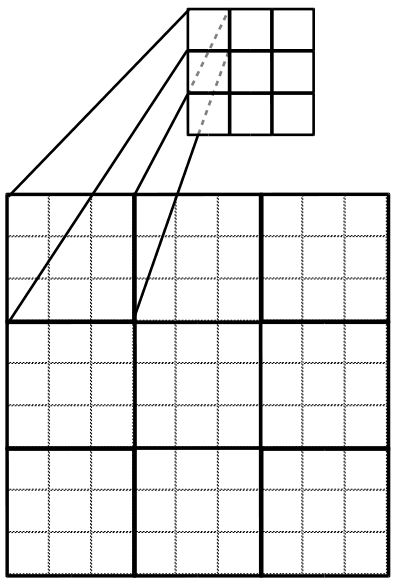}
  \hspace{1cm}
  \includegraphics[width=40mm]{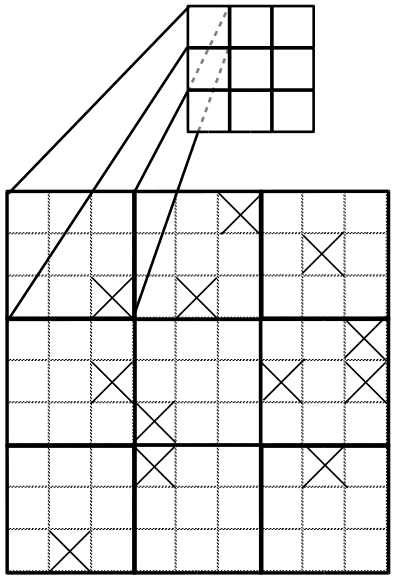}  
\caption{Free convolutional layers maintain a separate kernel at each location, unlike typical convolutional layers, that apply the same filter across all possible locations. The above figures are examples of FCN layers on a 9x9 input space. Each 3x3 subregion of the input is covered with a distinct kernel (weight matrix), as shown in the diagram on the left. The top square represents the output obtained from applying the filter to the corresponding input region. The diagram on the right depicts free convolutional layers with variable connection patterns, where the x's represent absent connections. In this example, 12 out of the 91 connections are missing, creating a variable connection probability of roughly 0.15.}
\label{fig:lc_visualization}
\end{figure}

Relaxing the weight-sharing assumption in CNNs yields a Free Convolutional Network (FCN). In FCNs, the weights of a filter at a specific location are not tied to the weights of the same filter at a different location (see Figure \ref{fig:lc_visualization}). This naturally means that FCNs have far more parameters than the corresponding CNN and are slower to train on a digital machine. However, this is not a concern here, as our primary goal is not to improve the efficiency of CNNs deployed on digital machines, but rather to understand the consequences of relaxing the weight-sharing assumption inside a native neural machine. 

Furthermore, it is highly implausible that a given neuron will share the same dendritic tree with a neighboring neuron \cite{Jan_2010}, thus in addition to neighboring neurons of the same layer not having the same weights, we would like to consider also the possibility of them not having the same receptive field pattern. Thus, in addition to plain FCNs, FCNs with variable connection patterns are implemented in the simulations below. The implementation of variable connection patterns has many options. Here, for simplicity, a random percentage of connections are severed (Figure \ref{fig:lc_visualization}) [Note: this is very different from dropout where different sets of weights get randomly set to 0 at each presentation of a training example]. Similar to in methodology to DropConnect \cite{wan2013regularization}, however, the same weights remain 0 for all of training and testing. The x's in the right image of Figure \ref{fig:lc_visualization} correspond to missing synaptic connections between neurons that are set to zero and never trained.  

By running simulations comparing CNNs and FCNs (with and without variable connection patterns), we seek to answer the research questions laid out in the introduction. Appendix C contains complete details regarding simulations with variable connection patterns.

\section{Data and Methods}

\begin{figure}
\hspace{-20mm}
\includegraphics[width=200mm]{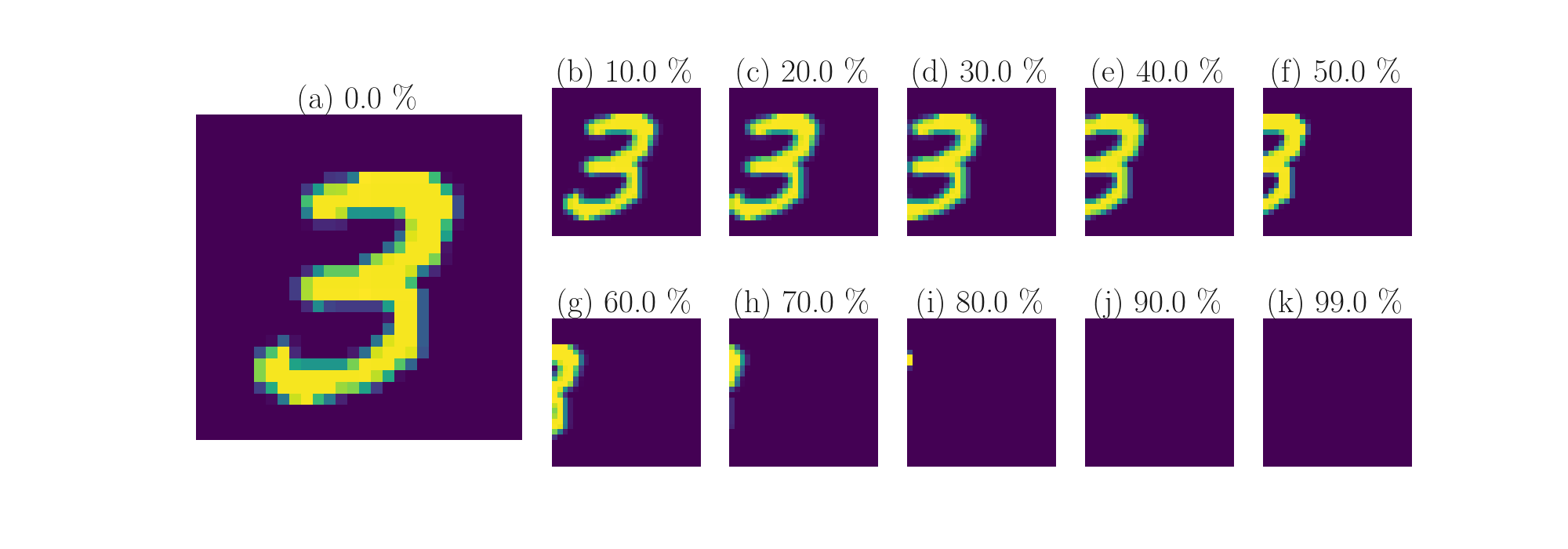}
\caption{Examples of translational augmentation on MNIST. In training images are translated left-right as well as up-down. The above only display the result of continually translating an image leftwards, for a visual example. (a) 0\% translation augmentation, equivalent to a un-altered MNIST image. (b-k) Gradually increasing the percentage of augmentation by 10\% each time. }
\label{fig:aug}
\end{figure}

In the simulations, we focus exclusively on computer vision tasks. We evaluate various free and shared weight networks on two well-known benchmark datasets: the handwritten digit dataset, MNIST \cite{lecun-mnisthandwrittendigit-2010}, as well as the CIFAR-10 object dataset \cite{cifar-10}. 

In the case of free weights, we consider using data augmentation by translating images horizontally and vertically to potentially compensate for the lack of translation, inherent in the architecture.
Due to the local receptivity of free weight networks, individual filters learn features solely within their receptive field. Translationally invariant data will allow filters to learn more or less the same features across the input space. Conversely, in CNNs, weights are shared across space. By applying the same operation across the input space, translational invariance is naturally embedded in the model. This property of CNNs gives them an inherent advantage in vision-based tasks. The purpose of these experiments is not to demonstrate the superiority of one network but to understand the consequences of weight-sharing and properties that arise around it.

Eleven settings of translation were tested in experiments (0\% to 99\% by increments of 10\%). Translational augmentation involves shifting images horizontally and vertically by varying degrees of the width and height respectively. Points outside the boundaries of the input get filled according to the nearest pixels. Figure \ref{fig:aug} shows examples of translational augmentation results on MNIST. During training, each image presented to the network is translated left-right as well as up-down by random amounts. 0\% to the augmentation setting for that trial. e.g., at 90\% augmentation, an image may be translated any amount 0-90\% up-down as well as 0-90\% left-right. 

All simulations, each augmentation setting, were completed using five-fold cross-validation. Simulations were implemented in Keras \cite{chollet2015keras} with a Tensorflow \cite{tensorflow2015-whitepaper} backend using NVIDIA GeForce GTX Titan X GPUs with 12 GB memory. For purposes of reproducibility, the code has been made publicly available\footnote{\url{https://github.com/jordanott/WeightSharing}}.

\subsection{Networks}
Conducting a grid search yielded appropriate hyperparameter configuration for all networks. See Appendix A.2 for complete details. The search produced network architectures composed of three convolutional/free convolutional layers, followed by a single hidden layer, and one output layer for classification.
A full description of all networks, including filter size, number of filters, stride, and learning rate can be found in Table A.2. MNIST and CIFAR require different architectures as the input sizes differ. For every setting of data augmentation (0 to 99\%, increments of 10\%) a CNN and FCN were trained via five-fold cross-validation. 

For simplicity, the architecture of CNNs and FCNs are equivalent, except that free convolutional layers replace convolutional layers. The activation functions, softmax layer, as well as the number of filters per layer,  remain the same across all networks. Table A.2 lists hyperparameter settings of the networks used in this paper. Additionally, it is essential to note that there is no architectural difference between the networks trained with data augmentation and those trained without.

\subsection{Variable Connection Patterns}
Also implemented in this study are neurons with variable connection patterns in FCNs. At the start of training, a chosen percentage of weights are randomly set to 0, representing the absence of a dendritic connection. These missing weights do not contribute to the output of the layer, and their values are never updated during backpropagation. The resulting connection patterns are maintained throughout training and testing. 
There are multiple options for implementing neurons with variable connection patterns. 
For computational simplicity, the implementation used in this paper is to turn off connections within each square filter given some probability (depicted in Figure \ref{fig:lc_visualization}). In simulations, we vary the drop probability from 0 to 99\% by increments of 10\%. The results from these simulations are reported in Figure C.1 and C.2 of the Appendix.

\section{Results}
\begin{figure}
\vspace{-10mm}
\hspace{-20mm}
\includegraphics[width=200mm]{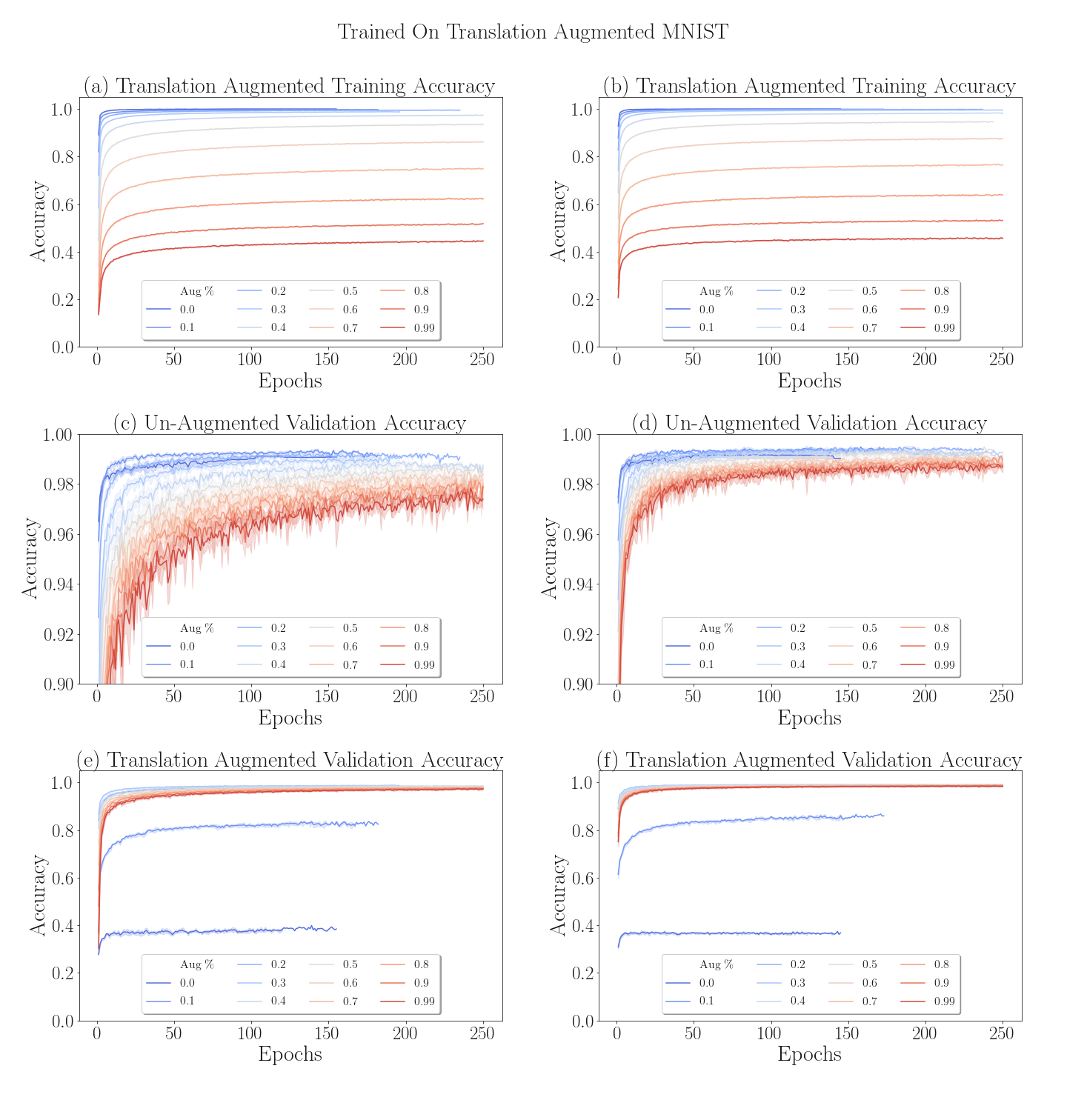}
\caption{MNIST results. Shown above are FCN (left column) and CNN (right column) results trained with varying degrees of translational augmentation, indicated by the legend. Top row: training accuracy over time. Middle row: validation accuracy over time. Bottom row: accuracy on the translationally augmented validation set. }
\label{fig:mnist}
\end{figure}

\begin{figure}
\vspace{-10mm}
\hspace{-20mm}
\includegraphics[width=200mm]{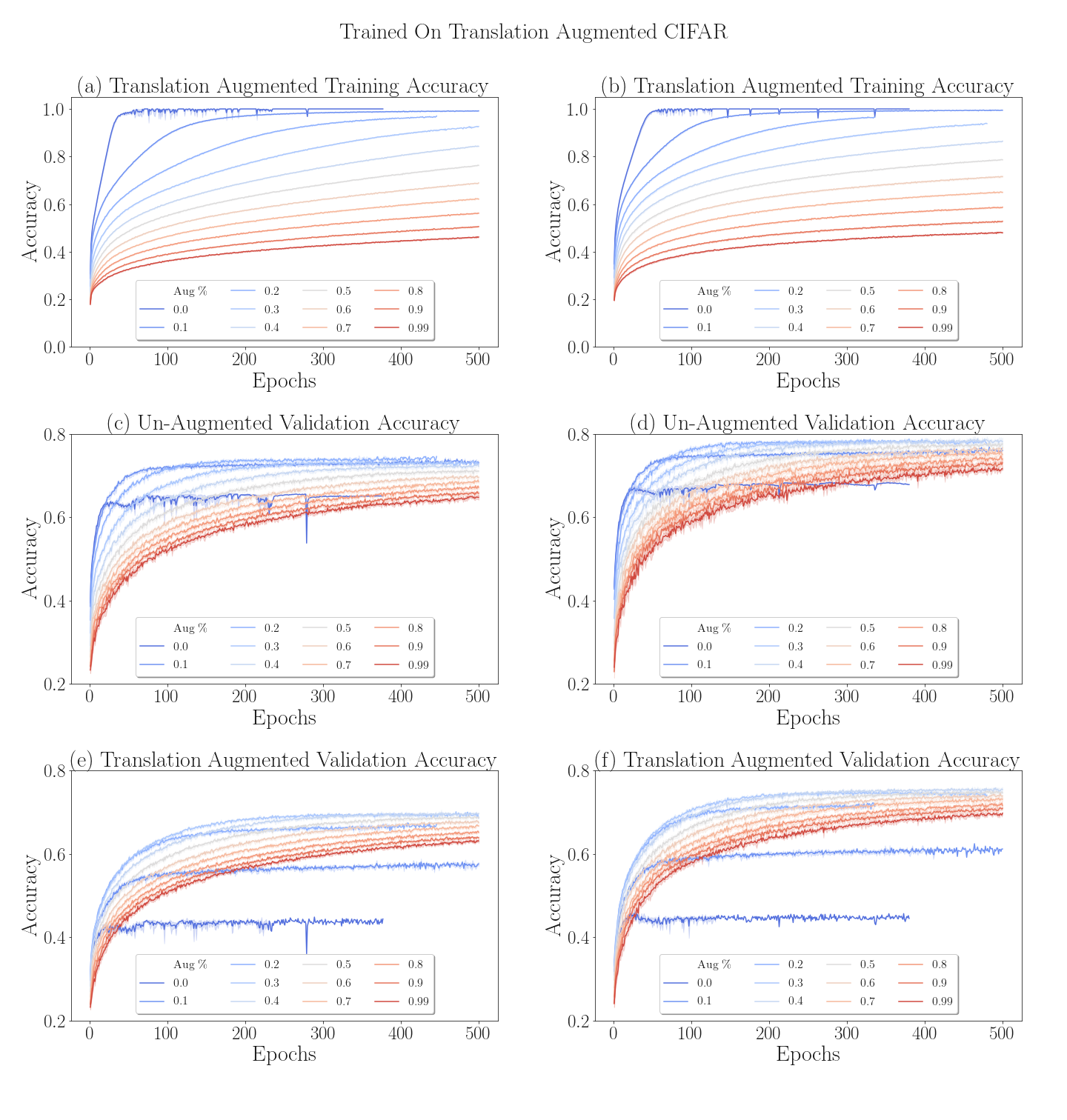}
\caption{CIFAR results. Shown above are FCN (left column) and CNN (right column) results trained with varying degrees of translational augmentation, indicated by the legend. Top row: training accuracy over time. Middle row: validation accuracy over time. Bottom row: accuracy on the translationally augmented validation set. }
\label{fig:cifar}
\end{figure}

We report accuracy metrics on translationally augmented training (0 - 99\%, increments of 10\%), un-augmented validation, and translationally augmented validation set (with 25\% translation) throughout training. Results for MNIST and CIFAR are shown in Figure \ref{fig:mnist} and \ref{fig:cifar}, respectively. The legend denotes the amount of translation used during training (0 - 99\%, increments of 10\%). 

Tables \ref{tab:mnist_results} and \ref{tab:cifar_results} show the median accuracy of the five-fold cross validation experiments for the MNIST and CIFAR-10 datasets respectively. The tables display results for each corresponding setting of translation augmentation (0 - 99\%, increments of 10\%). Bold values indicate the highest performing model in that accuracy metric for CNN and FCN, respectively. The Appendix contains additional experiments regarding the use of other augmentation methods, additional training metrics, as well as simulations regarding neurons with variable connection patterns.

\begin{table}[]
    \centering
    \begin{tabular}{l||cc|cc|cc}
        \toprule
        {} & \multicolumn{2}{c|}{Training Accuracy} & \multicolumn{2}{c|}{Validation Accuracy} & \multicolumn{2}{c}{Translation Accuracy} \\
        Aug \% &  CNN &  FCN & CNN & FCN & CNN &  FCN \\
        \midrule
        \midrule
        0.00 &  \textbf{0.999964} &  \textbf{0.999175} &  0.990536 &  0.988071 &    0.366536 &  0.372287 \\
        0.10 &  0.997793 &  0.995665 &  \textbf{0.993071} &  \textbf{0.991857} &    0.837963 &  0.813368 \\
        0.20 &  0.996719 &  0.991730 &  0.993000 &  0.990357 &    0.986220 &  0.977033 \\
        0.30 &  0.992555 &  0.983998 &  0.992143 &  0.987643 &    \textbf{0.990741} &  \textbf{0.985062} \\
        0.40 &  0.976966 &  0.965521 &  0.991571 &  0.984500 &    0.989439 &  0.982422 \\
        0.50 &  0.935973 &  0.923703 &  0.990071 &  0.981571 &    0.987594 &  0.979456 \\
        0.60 &  0.863680 &  0.847367 &  0.988286 &  0.978500 &    0.986111 &  0.976273 \\
        0.70 &  0.752630 &  0.731917 &  0.987000 &  0.974286 &    0.984484 &  0.971933 \\
        0.80 &  0.626956 &  0.606541 &  0.985500 &  0.970786 &    0.982820 &  0.968171 \\
        0.90 &  0.520896 &  0.503312 &  0.984714 &  0.967821 &    0.982096 &  0.965133 \\
        0.99 &  0.448491 &  0.431218 &  0.984000 &  0.965071 &    0.981120 &  0.961372 \\
        \bottomrule
    \end{tabular}
    \caption{MNIST results. Median accuracies of training, un-augmented validation, and augmented validation set. The left most column denotes the amount of translational augmentation used during training. When performing translational augmentation on the validation set, 25\% augmentation was used throughout the experiments. Bold values indicate the highest performing model in that accuracy metric for CNN and FCN, respectively.}
    \label{tab:mnist_results}
\end{table}

\begin{table}[]
    \centering
    \begin{tabular}{l||cc|cc|cc}
        \toprule
        {} & \multicolumn{2}{c|}{Training Accuracy} & \multicolumn{2}{c|}{Validation Accuracy} & \multicolumn{2}{c}{Translation Accuracy} \\
        Aug \% &  CNN &  FCN & CNN & FCN & CNN &  FCN \\
        \midrule
        \midrule
        0.00 &  \textbf{1.000000} &  \textbf{1.000000} &  0.671083 &  0.647333 &    0.444556 &  0.433468 \\
        0.10 &  0.982156 &  0.973354 &  0.749333 &  0.724333 &    0.598538 &  0.561324 \\
        0.20 &  0.876542 &  0.858094 &  0.773250 &  \textbf{0.731500} &    0.705225 &  0.653478 \\
        0.30 &  0.834646 &  0.781885 &  \textbf{0.774083} &  0.720167 &    0.739163 &  \textbf{0.683972} \\
        0.40 &  0.775188 &  0.725198 &  0.768750 &  0.704958 &    \textbf{0.742608} &  0.680444 \\
        0.50 &  0.712146 &  0.657438 &  0.751667 &  0.682250 &    0.729419 &  0.661458 \\
        0.60 &  0.650719 &  0.598344 &  0.734042 &  0.659958 &    0.714130 &  0.642557 \\
        0.70 &  0.592385 &  0.545354 &  0.717250 &  0.641542 &    0.699555 &  0.624286 \\
        0.80 &  0.535958 &  0.496677 &  0.699667 &  0.626250 &    0.681452 &  0.607695 \\
        0.90 &  0.483635 &  0.449490 &  0.684833 &  0.611667 &    0.667801 &  0.593834 \\
        0.99 &  0.442260 &  0.413802 &  0.668667 &  0.599458 &    0.653436 &  0.583375 \\
        \bottomrule
    \end{tabular}
    \caption{CIFAR results. Median accuracies of training, un-augmented validation, and augmented validation set. The left most column denotes the amount of translational augmentation used during training. When performing translational augmentation on the validation set, 25\% augmentation was used throughout the experiments. Bold values indicate the highest performing model in that accuracy metric for CNN and FCN, respectively.}
    \label{tab:cifar_results}
\end{table}

\begin{figure}
\hspace{-20mm}
\includegraphics[width=200mm]{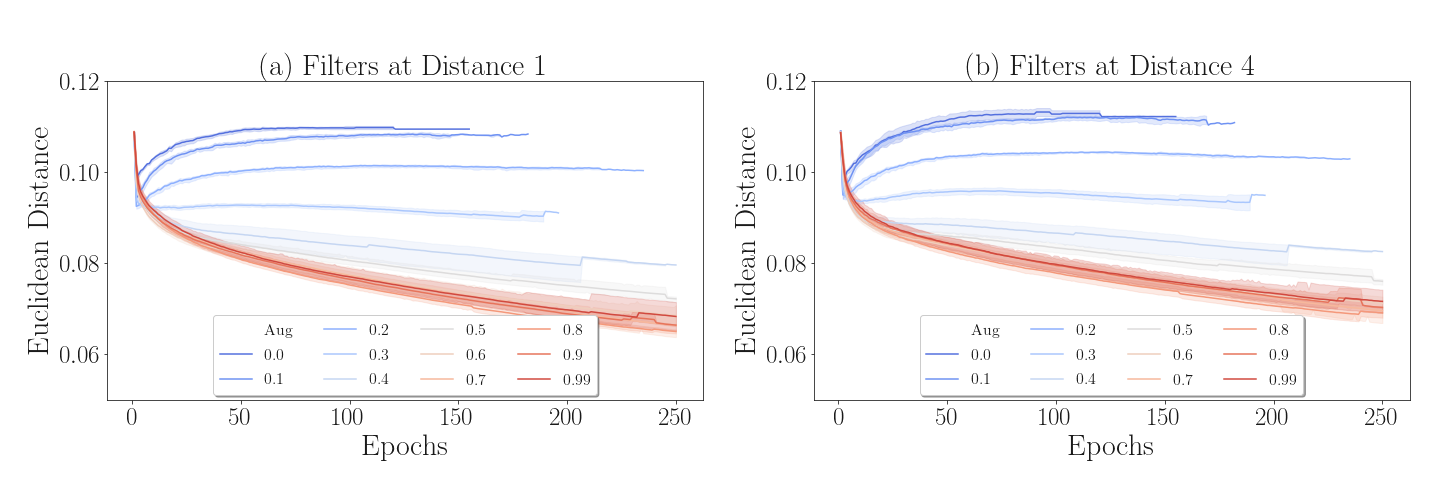}
\caption{Shown above is the average Euclidean distance between weight kernels in FCN layer overtime at the specified radius away. E.g., a radius of four indicates comparing a given weight kernel to all kernels four units away. As translational augmentation is increased weights become more similar to one another. As expected, weights closer in proximity (radius 4; left figure) have a smaller average distance than weights farther away (radius 7; right figure). }
\label{fig:approx_ws_translation}
\end{figure}

\begin{figure}
    \centering
    \includegraphics[width=100mm]{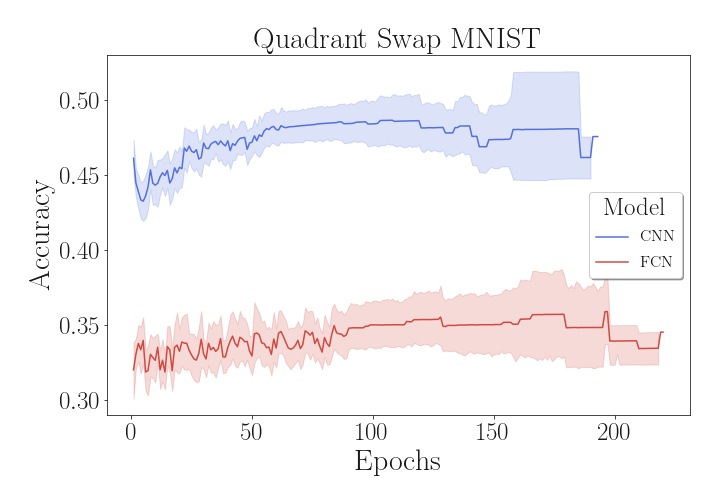}
    \caption{Results for CNN and FCN models trained on MNIST and tested on images where quadrant \rom{1} is replaced with quadrant \rom{3} and vice versa.}
    \label{fig:swap_results}
\end{figure}

\subsection{Is weight-sharing necessary to prevent overfitting?}
Both FCNs and CNNs are shown to overfit the training set, given a sufficient number of epochs. This result is evident from a divergence in training accuracy vs. validation accuracy over time (MNIST and CIFAR, Figures \ref{fig:mnist} and \ref{fig:cifar} respectively). In situations, without overfitting, one would expect training and validation scores to be close to one another. However, as the training accuracy continues to increase towards 100\%, the validation accuracy plateaus or declines, a visible indication of overfitting. In experiments performed on MNIST and CIFAR overfitting was observed for CNNs and FCNs when translation was not used, Figure \ref{fig:mnist} and \ref{fig:cifar} respectively. This effect manifests in CIFAR (Table \ref{tab:cifar_results}) where there is more than a 30\% gap between training accuracy and validation accuracy for both CNN and FCN, with no translational augmentation (i.e., 0\% translation).

In the case of MNIST, we see from Table \ref{tab:mnist_results} that FCNs, trained on augmented data (10\% translation), achieve a median accuracy of 99.1857\% on the validation set. Only slightly less than the 99.3071\% median accuracy achieved by standard CNNs.
On the more complicated CIFAR dataset, with 10\% translation, FCNs can come within 2\% of CNN accuracy on the validation set (Table \ref{tab:cifar_results}). Thus, we can infer that data augmentation alone is sufficient to prevent overfitting, and weight-sharing need not be leveraged to achieve this.

Using augmentation during training (Figure \ref{fig:mnist} and \ref{fig:cifar}) not only reduces overfitting but also leads to an increase in performance on validation sets. For FCNs on CIFAR specifically, training with translational amounts of 10, 20, and 30\% increases FCN validation performance while dramatically reducing overfitting, compared to 0\% translation. In human learning, translational augmentation comes as a byproduct of interacting with a changing world, which inherently provides the brain with samples of the same object translated at different positions - for example, a car driving down the street. Because FCNs cannot rely on weight-sharing to simulate the effect of translation, they must, therefore, rely on manual augmentation. \textit{Large, translationally invariant datasets, like those produced through data augmentation, are essential for free weight networks to achieve excellent performance and avoid overfitting. If this constraint can be met, overfitting can be mitigated without weight-sharing.}

\subsection{Is weight-sharing necessary to ensure translational invariant recognition?}
Learning translationally invariant representations is tested by using translational augmentation on the validation set. Figures \ref{fig:mnist}e and \ref{fig:mnist}f for MNIST and Figures \ref{fig:cifar}e and \ref{fig:cifar}f for CIFAR show results of translation on the augmented validation set. 

Significant disparity between validation accuracy and translation augmented validation accuracy would signal that the network is not capable of translationally invariant recognition. For example, the FCN trained on MNIST data without augmentation saw more than a sixty percent gap between validation set accuracy and translation augmented validation set accuracy (Table \ref{tab:mnist_results}). Similarly, on CIFAR, the FCN trained without translation augmentation performed 20\% worse on the translation augmented validation set than the standard one (Table \ref{tab:cifar_results}). Meaning these networks trained with 0\% augmentation are incapable of learning translation invariant representations. 

On both MNIST and CIFAR, FCNs achieve comparable results on the validation set and translation augmented validation set. This result is evident on MNIST where an FCN trained with 30\% translation achieves a 0.2\% difference between validation and translation augmented validation performance. Likewise, on CIFAR, the FCN trained with 40\% translation comes within 2\% of its validation performance on the translation augmented validation set. \textit{The results show that as translational augmentation is used during training, the gap between the validation accuracy and translation augmented validation accuracy decreases. This confirms that FCNs are capable of learning translationally invariant representations provided sufficient, translationally augmented, data.}

\subsection{Can acceptable performance be achieved without weight-sharing?}
FCNs can achieve high-performance scores on two benchmark computer vision datasets.
In the absence of data augmentation (0\% translation), CNNs outperform FCNs on MNIST and CIFAR. Referencing Table \ref{tab:mnist_results}, CNNs achieve 99.05\% validation accuracy whereas FCNs are slightly worse with 98.81\% validation accuracy for MNIST. On CIFAR, Table \ref{tab:cifar_results}, CNNs outperform FCNs with 67.11\% and 64.73\%, respectively. When FCNs are exposed to translational data, they can achieve 99.19\%, 10\% augmentation, and 73.15\%, 20\% augmentation, on MNIST and CIFAR validation sets, respectively. 

\textit{The validation set accuracy observed in our simulations shows that as translational augmentation is increased, FCNs can match the performance of CNNs. Thus, there is not a fundamental necessity of weight-sharing to attain satisfactory performance.}

\subsection{Does approximate or exact weight-sharing emerge in a natural way?}
To test for the emergence of approximate weight-sharing, we calculate the average euclidean distance between each filter of the FCN layer, at a specified radius. Experiments in this paper use a radius of one and four units to test the similarity of filters at different distances. Figure B.5 in the Appendix, gives a visual example. This depiction means that each filter in the FCN layer is compared to the filters one and four units, respectively, away from it. Figure \ref{fig:approx_ws_translation} reports the euclidean distance between filters at radius one and radius four. 

If weight-sharing did emerge naturally, one would expect to see a lower average euclidean distance between filters within an FCN layer throughout training. One would only expect to see this convergence in weight values if a similar stimulus is present across the input space (i.e., translationally augmented data, akin to video). Simulations conducted on the MNIST dataset confirms this. As the amount of translation increases (indicated by the legend in Figure \ref{fig:approx_ws_translation}), the parameters of the FCN layer converge to similar values; the euclidean distance between filters decreases. Furthermore, a higher degree of translation yields more similar the weight values. Conversely, when translation is not used during training, the weights diverge and become less similar from each other over time (i.e., the distance between them increases). 

One would also expect filters near one another to be more similar than those farther away (i.e., filters at radius 1 should be more similar than at radius 4). Again, this is confirmed in simulations where Figure \ref{fig:approx_ws_translation}a shows lower average Euclidean distance values, per augmentation setting, compared to Figure \ref{fig:approx_ws_translation}b. 

To ascertain the cause of approximate weight-sharing, FCNs were trained with noise and rotation augmentation while measuring the distance between their filters. Figure B.6 and B.7 display euclidean distance results for these experiments. These metrics confirm that only translation augmentation is sufficient to endow FCNs with approximate weight-sharing. Additional explanations are provided in Appendix B.5.

\textit{While not exact, the distance between FCN filters shows how approximate weight-sharing can emerge in a natural way with translationally augmented data.}

\subsection{For What Learning Tasks Are Free Convolutional Networks Most Applicable?}
In CNNs, the same filter is applied to all locations in the input space. One may hypothesize that, as a result, CNNs will be focused less on the overall structure of features in spatial relation to each other, and more on identifying the features themselves. This is in contrast to FCNs, which can only operate locally within their receptive field, and so must inherently learn the global structure of features in relation to one another. 

To test this hypothesis, images presented to the networks for testing need to be manipulated in such a way so as to compromise the overall global structure of the image while at the same time preserving individual features appearing within an image. Appendix B.3 details the quadrant swap task, in which quadrants \rom{1} and \rom{3} of images are interchanged with each other in the validation set. See Figure B.3 for a visual example. In this task, lower classification accuracy would indicate a higher importance placed on global structure during training. That is, it is not sufficient for learned features just to be present in an image, but the spatial relationship between them must also be preserved to some degree.

The results from the quadrant swap task are shown in Figure \ref{fig:swap_results}. One might expect severe degradation of performance after rearranging parts on an image. However, the CNN performs at nearly 50\% accuracy. This performance indicates that the CNN is still able to recognize features of the altered images when making its prediction. Conversely, the FCN performs nearly 20\% worse. Higher accuracy bolsters the notion that the overall global structure of an image is less important for a CNN as opposed to the FCN. 
\textit{In domains where global structure is essential such as face recognition or biomedical imaging, FCNs may find substantial applicability.}

\subsection{Is weight-sharing necessary?}
The necessity of weight-sharing arises as a means of parameter reduction. Reducing the number of parameters leads to networks that are faster to train and smaller to store. Due to the limitations of modern hardware, practical applications of free weight networks are not currently viable, especially in embedded environments. Thus in situations where space is a constraint, and translationally invariant datasets are not available, weight-sharing becomes a necessity. 

\textit{In terms of accuracy, FCNs have shown comparable results are possible without weight-sharing. In this regard, weight-sharing is not a necessity.}

\subsection{If weight-sharing is not necessary, are translational invariant training sets necessary?}
To examine the necessity of translational datasets, additional experiments were conducted using noise and rotational augmentation during training. Intuition would suggest that only translational data is sufficient to endow FCNs with translational invariant recognition.
The full results, including accuracy performance on the noise and rotation, augmented training, un-augmented validation, rotation augmented validation, noise augmented validation, and translation augmented validation set can be found in Appendix B. 

Training using other augmentation methods that do not produce translational invariant datasets yield poor results when testing on the translationally augmented validation set. Referencing Appendix B (Table B.1, B.2, and B.3), the results show that FCNs trained on non-translational data are consistently not capable of learning translationally invariant representations. Low performance indicates that only translationally augmented training data allows FCNs to learn translationally invariant representations. 
\textit{Using translationally invariant datasets yields better results in validation set and augmented validation set accuracy, specifically in FCNs.}

\section{Conclusion}
The use of weight-sharing arises as a solution to parameter reduction and translationally invariant recognition in neural networks. Though weight-sharing is implausible in any biological or physical setting, it is instrumental in computer vision tasks. We have examined alternatives to weight-sharing, such as free convolutional networks, where the weight-sharing assumption is relaxed. FCNs trained with augmented datasets have been shown to match and even surpass standard CNNs in validation set accuracy. Data augmentation, specifically datasets augmented via translation, is a necessity as a means to avoid overfitting and train FCNs capable of translationally invariant recognition. Thus, in environments where data is plentiful and computational resources can cope with the large number of parameters that result from abandoning weight-sharing, FCNs provide an alternative to CNNs that can achieve potentially superior performance and higher fidelity to physical systems.

\appendix
\section*{A. Experimental Settings}

\subsection*{A.1. Hyperparameter Search}
\begin{table}[]
\footnotesize
\centering
\label{hyperparameter-table}
\begin{tabular}{@{}ll@{}ll@{}}
\toprule
Name                        & Options    & Parameter Type         \\
\midrule
Learning Rate               & $10^{-i},$ $i \in [1,6]$ \hspace{20mm}& Choice \\
Number of layers            & {[}2, 3{]}        \hspace{20mm} &   Discrete \\
\bottomrule
\end{tabular}
\caption*{Table A.1 Hyperparameter Space for the conducted grid search.}
\end{table}

We conducted a hyperparameter search to explore the space of possible architectures. CNN and FCNs trained on MNIST and CIFAR, using a grid search to find the optimal setting. The search was executed using SHERPA \citep{hertel2018sherpa}, a Python library for hyperparameter tuning. We detail the hyperparameters of interest in Table A.1, as well as the range of available options during the search.

During the hyperparameter search, MNIST trained for 100 epochs with a patience of 25 monitoring the validation accuracy. CIFAR trained for 200 epochs with a patience of 50 monitoring the validation accuracy. We show the hyperparameters of the best-performing networks in Table A.2. All networks achieved the best performance with three layers. The ultimate difference between MNIST and CIFAR architectures is the learning rate, $10^{-3}$ and $10^{-4}$, respectively. 

\subsection*{A.2. Network Architectures}
Table A.2 describes the architectures used in this paper. MNIST networks have three convolutional or free convolutional layers respective of the architecture. All weight kernels are of size 3x3 with 32, 64, and 128 filters for the three layers. This output feeds into a fully connected layer of 1024 nodes, followed by a softmax layer for classification.

CIFAR networks have three convolutional or free convolutional layers respective of the architecture. Layer one has a 5x5 weight kernel, 64 filters, and a stride of 2. Layer two has a 5x5 weight kernels, 128 filters, and a stride of 2. Layer three has 3x3 weight kernels, 256 filters, and a stride of 1. Following these layers are a fully connected layer of 1024 nodes, followed by a softmax layer for classification.

\begin{table}[]
    \centering
    \begin{tabular}{c|c|c}
          & CNN & FCN  \\
         \midrule
         \multirow{7}{*}{MNIST} & Convolutional(3x3,32,2) & Free Convolutional(3x3,32,2) \\
          & Convolutional(3x3,64,2) & Free Convolutional(3x3,64,2) \\
          & Convolutional(3x3,128,1) & Free Convolutional(3x3,128,1) \\
          & Fully Connected(1024) & Fully Connected(1024) \\
          & Softmax(10) & Softmax(10) \\
          & & \\
          & Learning Rate: $10^{-3}$ & Learning Rate: $10^{-3}$ \\
          \midrule
          \multirow{7}{*}{CIFAR} & Convolutional(5x5,64,2) & Free Convolutional(5x5,64,2) \\
          & Convolutional(5x5,128,2) & Free Convolutional(5x5,128,2) \\
          & Convolutional(3x3,256,1) & Free Convolutional(3x3,256,1) \\
          & Fully Connected(1024) & Fully Connected(1024) \\
          & Softmax(10) & Softmax(10) \\
          & & \\
          & Learning Rate: $10^{-4}$ & Learning Rate: $10^{-4}$ \\
    \end{tabular}
    \caption*{Table A.2 Architecture specification resulting from the grid search. The format for convolutional layers is (kernel size, number of output channels, stride). All layers, except the output, use ReLU activations.}
    \label{tab:my_label}
\end{table}

\subsection*{A.3. Implementation Details}
Random seeds were set to zero for the Tensorflow backend and Numpy.
Batch sizes were 256 and 128 for MNIST and CIFAR, respectively. 
Weights for all layers were initialized using the Xavier Uniform Initialization. The source code for all experiments has been made publicly available at: \url{https://github.com/jordanott/WeightSharing}

\section*{B. Other Augmentation Methods}
To examine the necessity of translational datasets, additional experiments were conducted using noise and rotational augmentation during training. The hypothesis being that only translational data is sufficient to endow FCNs with translational invariant recognition.
The full results for edge noise, noise, and rotation augmentation can be found in Tables B.1, B.2, and B.3, respectively. 

Training using other augmentation methods that do not produce translational invariant datasets yield poor results when testing on the translationally augmented validation set. This result indicates that only translationally augmented training data allows FCNs to learn translationally invariant representations. 
\textit{Using translationally invariant datasets yields better results in validation set and augmented validation set accuracy, specifically in FCNs.}

\subsection*{B.1. Edge Noise}
Augmentation with edge noise represents adding Gaussian noise to the periphery of an image. For experiments in this paper, the periphery is defined as the 5 pixels bordering an image (for both CIFAR and MNIST). This type of augmentation serves to test the network's resiliency to noise on the fringe. For datasets like MNIST and CIFAR that contain the object of interest in center focus, this noise is unlikely to corrupt the object. Figure B.1 shows the effect of edge noise on the periphery, starting from zero noise, Figure B.1a, up to a standard deviation of 0.99, Figure B.1k.

\begin{figure}
    \hspace{-20mm}
    \includegraphics[width=200mm]{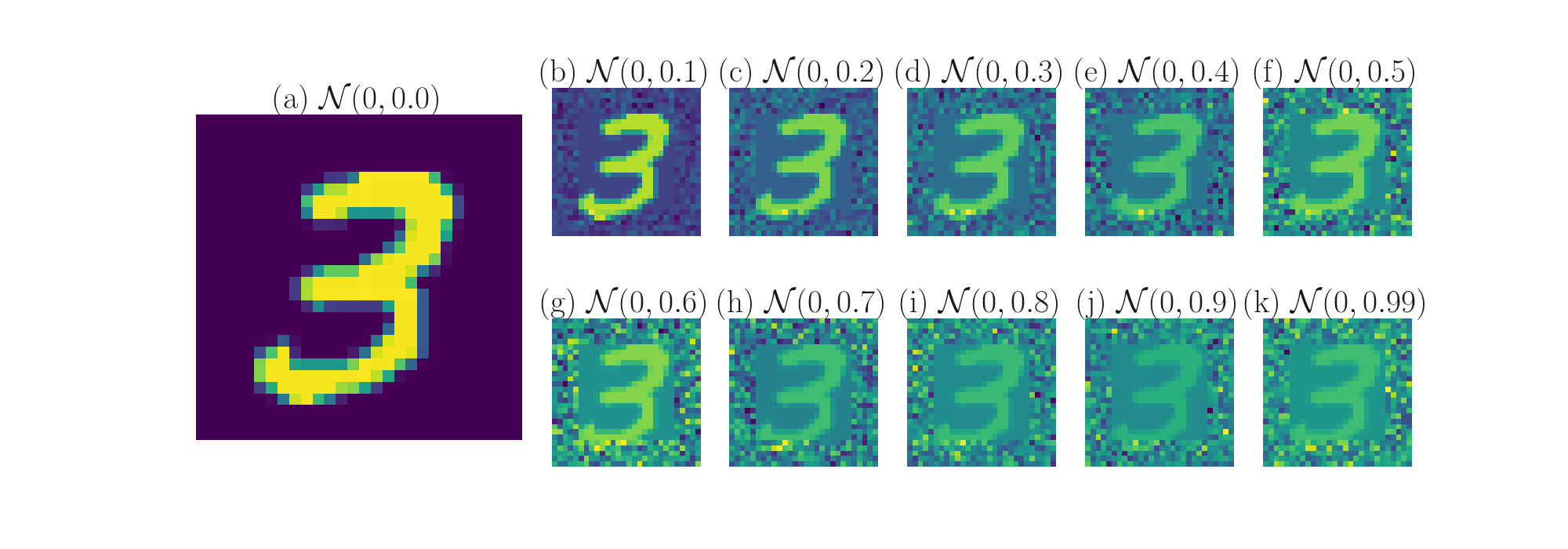}
    \caption*{B.1 Examples of edge noise augmentation on MNIST. (a) 0 variance noise, equivalent to a un-altered MNIST image. (b-k) Gradually increasing the variance of noise augmentation by .1 each time.}
    \label{fig:aug_edge_noise}
\end{figure}

\begin{table}[]
    \hspace{-7mm}
    \begin{tabular}{l||cc|cc|cc|cc}
    \toprule
    {} &  \multicolumn{4}{c|}{MNIST} & \multicolumn{4}{c}{CIFAR} \\
    {} &  \multicolumn{2}{c|}{Validation Acc} & \multicolumn{2}{c|}{Translation Acc} & \multicolumn{2}{c|}{Validation Acc} & \multicolumn{2}{c}{Translation Acc} \\
    Aug \% & CNN & FCN & CNN & FCN & CNN & FCN & CNN & FCN \\
    \midrule \midrule
    0.00 &  0.99054 &  \textbf{0.98807} &     \textbf{0.36654} &  0.37229 &  \textbf{0.67108} &  \textbf{0.64733} &  \textbf{0.44456} &  \textbf{0.43347} \\
    0.10 &  \textbf{0.99100} &  0.98786 &     0.36263 &  \textbf{0.37666} &  0.64475 &  0.62633 &     0.42742 &  0.42179 \\
    0.20 &  0.98921 &  0.98714 &     0.35503 &  0.37587 &  0.63567 &  0.62375 &     0.42288 &  0.42011 \\
    0.30 &  0.98943 &  0.98714 &     0.35113 &  0.36777 &  0.62992 &  0.62350 &     0.41944 &  0.41986 \\
    0.40 &  0.98950 &  0.98750 &     0.34368 &  0.36053 &  0.61225 &  0.62342 &     0.40835 &  0.41952 \\
    0.50 &  0.98864 &  0.98664 &     0.34107 &  0.35236 &  0.60917 &  0.62167 &     0.40348 &  0.41734 \\
    0.60 &  0.98893 &  0.98750 &     0.33550 &  0.34968 &  0.59083 &  0.61617 &     0.39365 &  0.41541 \\
    0.70 &  0.98843 &  0.98707 &     0.33565 &  0.34650 &  0.59308 &  0.61558 &     0.39415 &  0.41473 \\
    0.80 &  0.98821 &  0.98671 &     0.33200 &  0.34165 &  0.58392 &  0.61375 &     0.38794 &  0.41322 \\
    0.90 &  0.98800 &  0.98679 &     0.33377 &  0.33970 &  0.58058 &  0.60925 &     0.38458 &  0.41112 \\
    0.99 &  0.98829 &  0.98650 &     0.32986 &  0.33767 &  0.57383 &  0.60996 &     0.37912 &  0.41053 \\
    \bottomrule
    \end{tabular}
    \caption*{B.1 Edge noise results. Median accuracy of un-augmented validation and translation augmented validation set. The left most column denotes the amount variance of noise used during training. When performing translational augmentation on the validation set, 25\% augmentation was used throughout the experiments. Bold values indicate the highest performing model in that accuracy metric for CNN and FCN, respectively.}
    \label{tab:edge_noise_results}
\end{table}

\subsection*{B.2. Noise}
Augmentation with noise represents adding Gaussian noise to the image. This type of augmentation serves to test the network's resiliency to noise corruption. Figure B.2 shows the effect of edge noise on the periphery, starting from zero noise, Figure B.2a, up to a standard deviation of 0.99, Figure B.2k.

\begin{figure}
    \hspace{-20mm}
    \includegraphics[width=200mm]{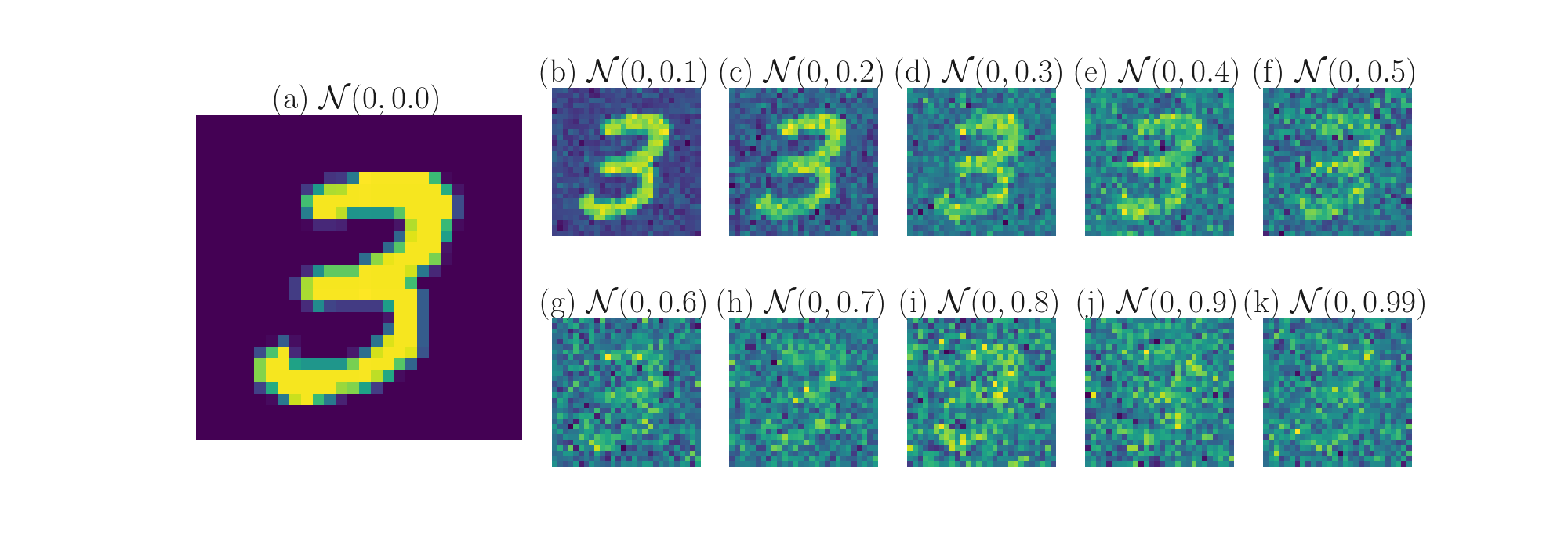}
    \caption*{B.2 Examples of noise augmentation on MNIST. (a) 0 variance noise, equivalent to a un-altered MNIST image. (b-k) Gradually increasing the variance of noise augmentation by .1 each time.}
    \label{fig:aug_noise}
\end{figure}

\begin{table}[]
    \hspace{-7mm}
    \begin{tabular}{l||cc|cc|cc|cc}
    \toprule
    {} &  \multicolumn{4}{c|}{MNIST} & \multicolumn{4}{c}{CIFAR} \\
    {} &  \multicolumn{2}{c|}{Validation Acc} & \multicolumn{2}{c|}{Translation Acc} & \multicolumn{2}{c|}{Validation Acc} & \multicolumn{2}{c}{Translation Acc} \\
    Aug \% & CNN & FCN & CNN & FCN & CNN & FCN & CNN & FCN \\
    \midrule \midrule
    0.00 &  \textbf{0.99054} &  0.98807 &     \textbf{0.36654} &  0.37229 & \textbf{0.67108} &  \textbf{0.64733} &     \textbf{0.44456} &  \textbf{0.43347} \\
    0.10 &  0.99000 &  0.98750 &     0.35757 &  0.37004 & 0.63092 &  0.63425 &     0.41507 &  0.41919 \\
    0.20 &  0.98957 &  0.98779 &     0.35880 &  \textbf{0.37410} & 0.59258 &  0.61825 &     0.39037 &  0.40381 \\
    0.30 &  0.99007 &  0.98829 &     0.36328 &  0.36957 & 0.56258 &  0.59404 &     0.37433 &  0.38899 \\
    0.40 &  0.98979 &  \textbf{0.98879} &     0.35793 &  0.36769 & 0.53521 &  0.56983 &     0.35652 &  0.37576 \\
    0.50 &  0.98964 &  0.98857 &     0.35084 &  0.34990 & 0.51142 &  0.55075 &     0.34232 &  0.36794 \\
    0.60 &  0.98900 &  0.98836 &     0.33623 &  0.33261 & 0.48083 &  0.53192 &     0.32451 &  0.35685 \\
    0.70 &  0.98807 &  0.98750 &     0.31854 &  0.31782 & 0.44858 &  0.50850 &     0.30612 &  0.34341 \\
    0.80 &  0.98614 &  0.98586 &     0.30433 &  0.30136 & 0.42417 &  0.47683 &     0.29255 &  0.32502 \\
    0.90 &  0.98407 &  0.98350 &     0.29080 &  0.28516 & 0.41179 &  0.44933 &     0.28642 &  0.30855 \\
    0.99 &  0.98164 &  0.98086 &     0.27590 &  0.27235 & 0.38250 &  0.40042 &     0.27050 &  0.28154 \\
    \bottomrule
    \end{tabular}
    \caption*{B.2 Noise results. Median accuracy of un-augmented validation and translation augmented validation set. The left most column denotes the amount variance of noise used during training. When performing translational augmentation on the validation set, 25\% augmentation was used throughout the experiments. Bold values indicate the highest performing model in that accuracy metric for CNN and FCN, respectively.}
    \label{tab:noise_results}
\end{table}

\subsection*{B.3. Quadrant Swap}
Using the convention from Cartesian geometry, quadrant \rom{1} and \rom{3} of images are swapped. This type of image deformation serves to test the networks reliance on features as opposed to global structure. The hypothesis being that because of CNNs shared weight paradigm, the location of features matters less opposed to the presence of the feature itself. Conversely, FCN filters can only operate locally within their receptive field, rendering the global structure of the image essential. 

To test this hypothesis, features of images need to be manipulated in such a way to compromise the overall structure of the image but preserve individual features. Figure B.3 shows a visual example of the swap procedure.

The results from this task, shown in Figure \ref{fig:swap_results}, indicate CNNs score a higher accuracy on Quadrant swapped images compared to FCNs. This indicates that CNNs are still able to recognize features of the altered images when making their prediction. This bolsters the notion that the overall structure of an image is less important for a CNN as opposed to the FCN, that performs nearly 20\% worse. In this task, a lower score indicates higher importance on the global structure.

\begin{figure}
    \hspace{-20mm}
    \includegraphics[width=200mm]{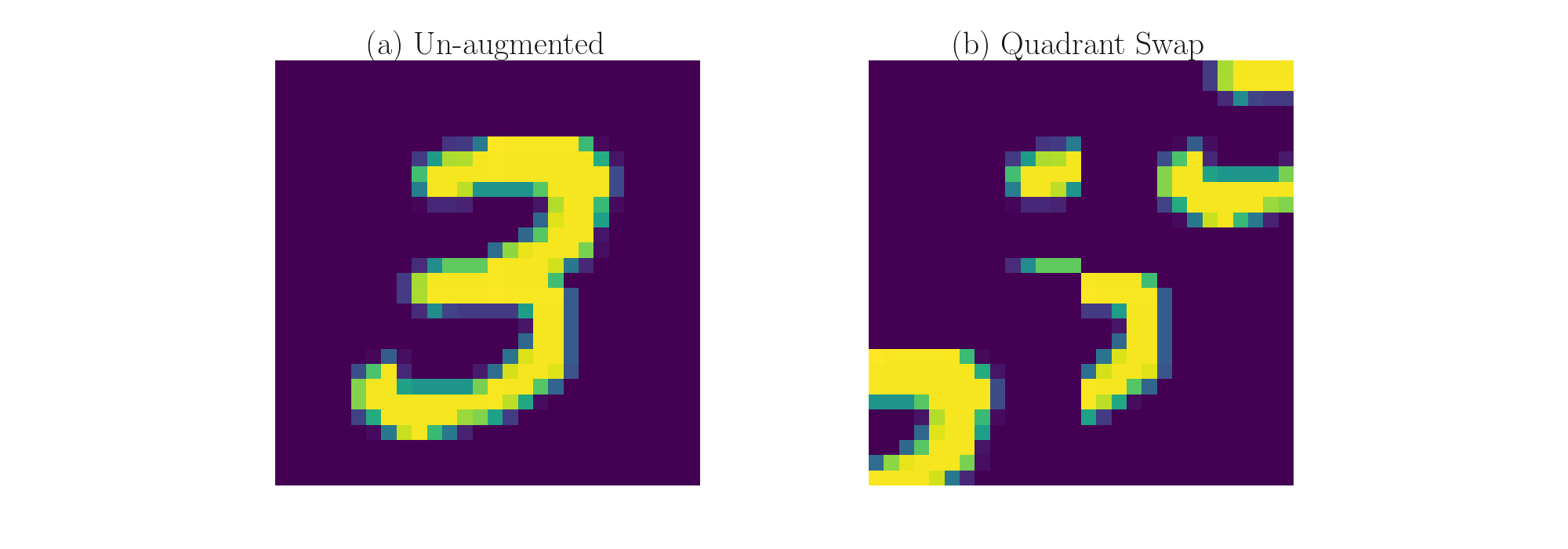}
    \caption*{B.3 Quadrant swap on MNIST. (a) Un-augmented image. (b) Quadrant swap augmentation. Where quadrant \rom{1} is replaced with quadrant \rom{3} and \rom{3} with \rom{1}}
    \label{fig:swap}
\end{figure}

\subsection*{B.4. Rotation}
Rotation augmentation is accomplished via rotating images clockwise about the center point. Figure B.4 shows the effect of rotating an MNIST image from 0\%, Figure B.4a, up through 99\%, Figure B.4k. Table B.3 displays the results of training networks with rotation augmentation.  

\begin{figure}
    \hspace{-20mm}
    \includegraphics[width=200mm]{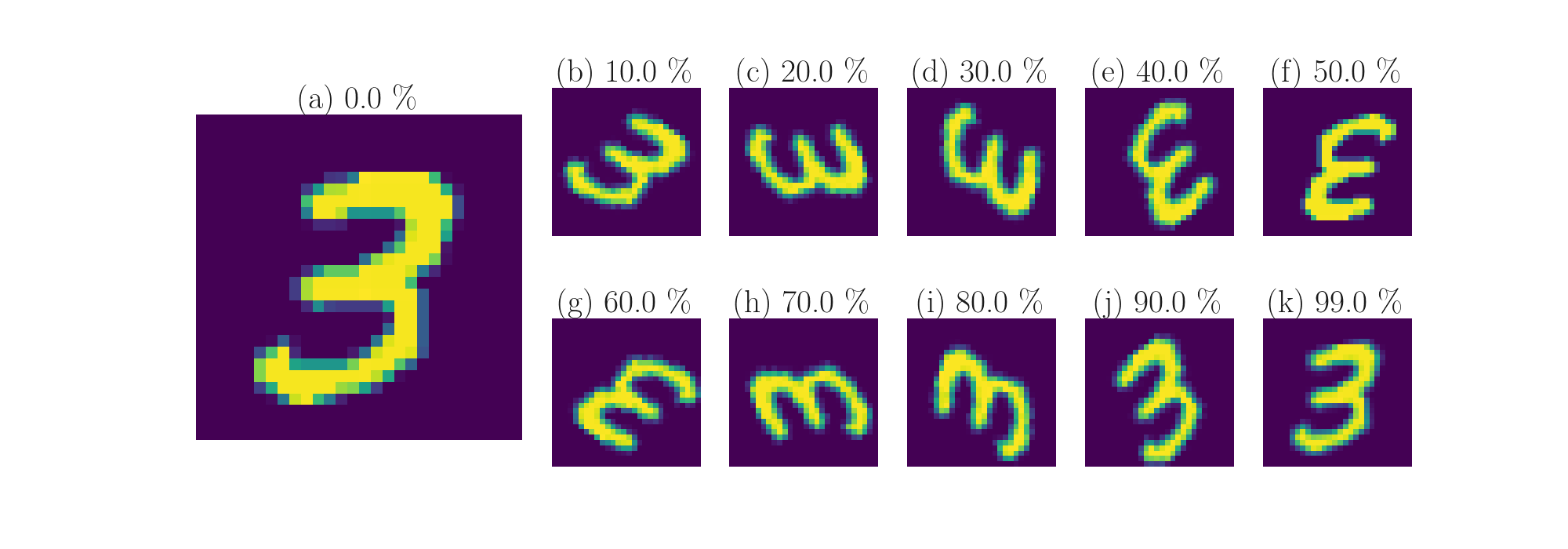}
    \caption*{B.4 Examples of rotation augmentation on MNIST. (a) 0\% rotation, equivalent to a un-altered MNIST image. (b-k) Gradually increasing the degree rotation by 10\% each time.}
    \label{fig:aug_rotation}
\end{figure}

\begin{table}[]
    \hspace{-7mm}
    \begin{tabular}{l||cc|cc|cc|cc}
    \toprule
    {} &  \multicolumn{4}{c|}{MNIST} & \multicolumn{4}{c}{CIFAR} \\
    {} &  \multicolumn{2}{c|}{Validation Acc} & \multicolumn{2}{c|}{Translation Acc} & \multicolumn{2}{c|}{Validation Acc} & \multicolumn{2}{c}{Translation Acc} \\
    Aug \% & CNN & FCN & CNN & FCN & CNN & FCN & CNN & FCN \\
    \midrule \midrule
    0.00 &  0.99054 &  0.98807 &     0.36654 &  0.37229 &  0.67108 &  0.64733 &     0.44456 &  0.43347 \\
    0.10 &  \textbf{0.99079} &  \textbf{0.98836} & \textbf{0.38527} &  \textbf{0.37753} & \textbf{0.68233} &  \textbf{0.65500} &  \textbf{0.48660} & \textbf{0.46211} \\
    0.20 &  0.98907 &  0.98571 &     0.36372 &  0.35055 &  0.65746 &  0.62608 &     0.47450 &  0.44766 \\
    0.30 &  0.98779 &  0.98436 &     0.33587 &  0.30107 &  0.64708 &  0.61000 &     0.47106 &  0.43364 \\
    0.40 &  0.98636 &  0.98329 &     0.32147 &  0.28877 &  0.63625 &  0.59767 &     0.46455 &  0.42981 \\
    0.50 &  0.98439 &  0.98114 &     0.32183 &  0.29492 &  0.62242 &  0.58550 &     0.46337 &  0.42465 \\
    0.60 &  0.98400 &  0.97979 &     0.32154 &  0.29854 &  0.61533 &  0.57458 &     0.45682 &  0.41919 \\
    0.70 &  0.98336 &  0.97864 &     0.32060 &  0.29337 &  0.60717 &  0.57100 &     0.44750 &  0.41465 \\
    0.80 &  0.98236 &  0.97786 &     0.31547 &  0.29015 &  0.60875 &  0.56567 &     0.44758 &  0.40961 \\
    0.90 &  0.98071 &  0.97686 &     0.31040 &  0.28566 &  0.59958 &  0.55767 &     0.44414 &  0.40692 \\
    0.99 &  0.98350 &  0.97975 &     0.32429 &  0.29239 &  0.61842 &  0.58108 &     0.46102 &  0.42221 \\
    \bottomrule
    \end{tabular}
    \caption*{B.3 Rotation results. Median accuracies of un-augmented validation and translation augmented validation set. The left most column denotes the percentage of rotation used during training. When performing translational augmentation on the validation set, 25\% augmentation was used throughout the experiments. Bold values indicate the highest performing model in that accuracy metric for CNN and FCN, respectively.}
    \label{tab:rotation_results}
\end{table}

\subsection*{B.5 Approximate Weight-Sharing}
The distance between weight kernels within an FCN layer was measured during training with other types of augmentation. This experiment tests if translation augmentation is the cause of approximate weight-sharing in FCNs. Figure B.6 and B.7 show the results of training with rotation and noise augmentation, respectively.  

The results show that for all augmentation settings of rotation, the distance between filters increases over time. 
Additionally, in noise augmented training, the average Euclidean distance increases in all cases except very high values of noise (0.8, 0.9, 0.99). Indicating this increase in filter similarity is due to the high noise levels across the image. This is confirmed visually by examining Figure B.2i, B.2j, and B.2k. Also, the decrease in euclidean distance for noise is not as dramatic as observed for translation, Figure \ref{fig:approx_ws_translation}. 

\begin{figure}
    \centering
    \begin{tabular}{cc}
        \includegraphics[width=.41\linewidth]{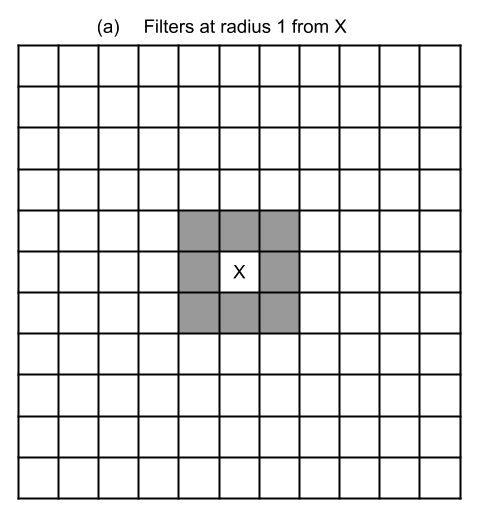} & \includegraphics[width=.4\linewidth]{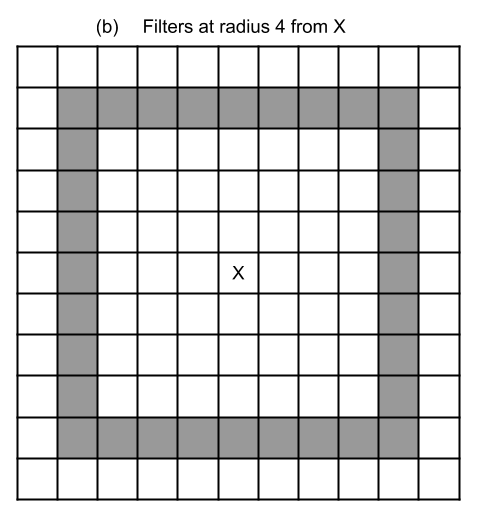} \\
    \end{tabular}
    \caption*{B.5 Filters at a specified radius. (a) Filters at radius 1 from the desired filter, marked by an X. (b) Filters at radius 4 from the desired filter, marked by an X.}
\end{figure}

\begin{figure}
    \hspace{-20mm}
    \includegraphics[width=200mm]{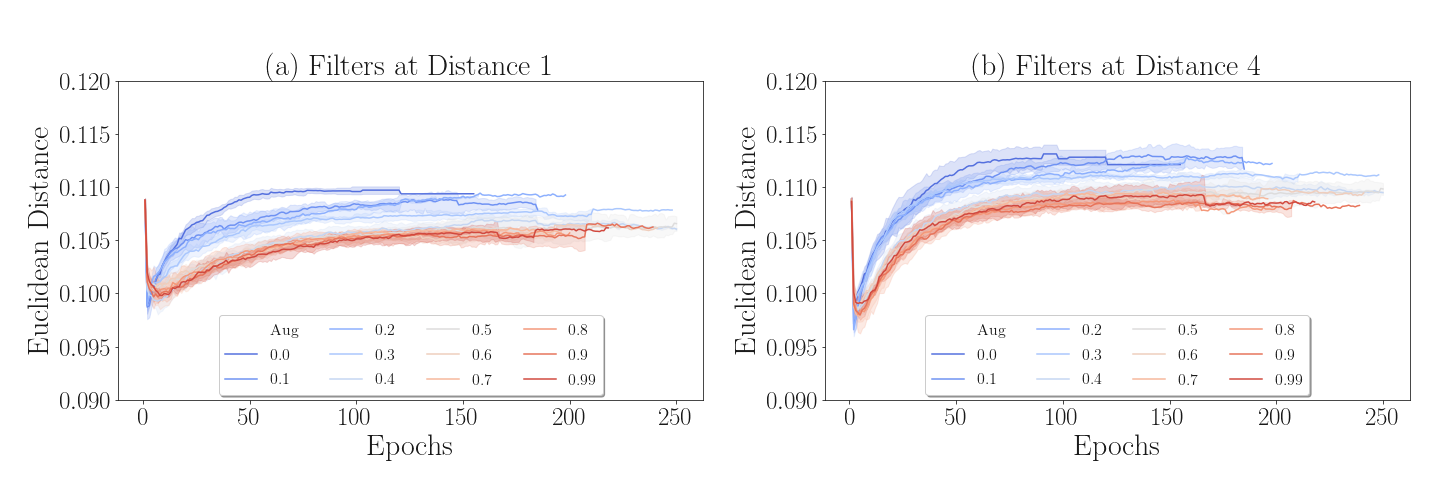} 
    \caption*{B.6 Euclidean distance between filters of an FCN layer, trained on MNIST with rotation augmentation. (a) Average euclidean distance between filters one unit away. i.e. all adjacent filters in the layer. (b) Average euclidean distance between filters four units away.}
    \label{fig:approx_ws_rotation}
\end{figure}

\begin{figure}
    \hspace{-20mm}
    \includegraphics[width=200mm]{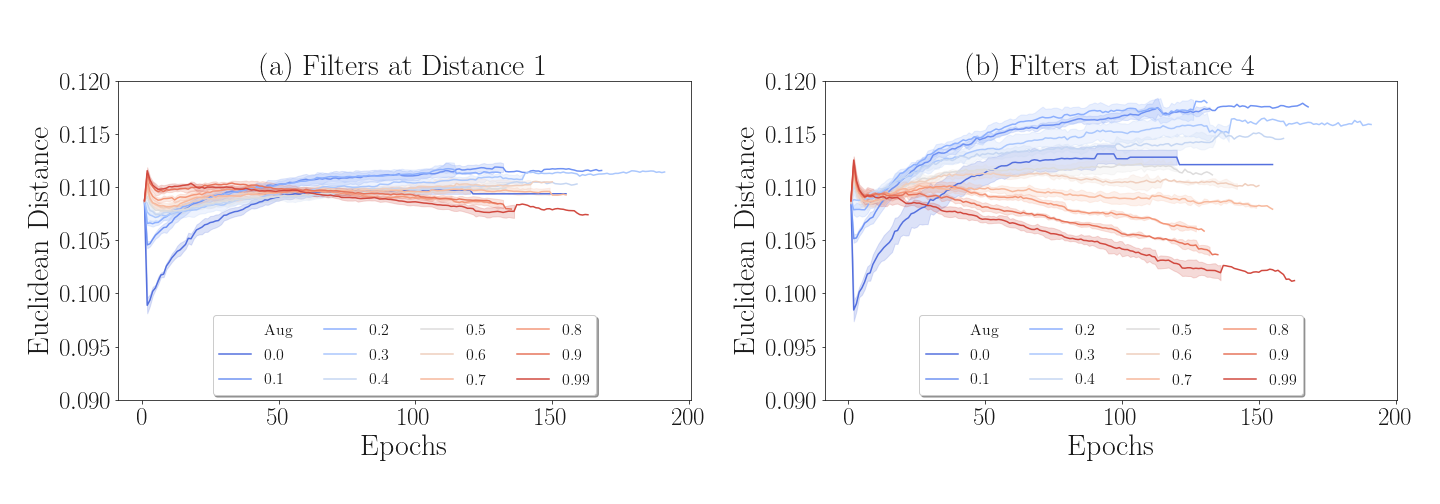}
    \caption*{B.7 Euclidean distance between filters of an FCN layer, trained on MNIST with noise augmentation. (a) Average euclidean distance between filters one unit away. i.e. all adjacent filters in the layer. (b) Average euclidean distance between filters four units away.}
    \label{fig:approx_ws_noise}
\end{figure}

\section*{C. Variable Connection Patterns}
Also implemented in this study are neurons with variable connection patterns in FCNs. At the start of training, a chosen percentage of weights are randomly set to 0, representing the absence of a dendritic connection. These missing weights do not contribute to the output of the layer, and their values are not updated during backpropagation. The resulting connection patterns are maintained throughout training and testing. 
There are multiple options for implementing neurons with variable connection patterns. 
For computational simplicity, the implementation used in this paper is to turn off connections within each square filter given some probability. In simulations, we vary the probability from 0 to 99\% by increments of 10\%. The results from these simulations are reported in Figure C.1 and C.2 for MNIST and CIFAR, respectively. Both datasets use 30\% translational augmentation for these experiments.

The variable connection probability is varied from 0\% to 99\% as indicated by the legend (VCP \%). The results shown in Figure C.1 and C.2 were trained using 20\% translation augmentation. Accuracy on the validation set indicates FCNs are robust to even large amounts of missing connections. Even when 80\% of connections are absent, the FCN is able to perform comparably well on both MNIST and CIFAR. FCNs are shown to be robust to a high degree of missing connections. Performance degrades rapidly beyond 90\%.

\begin{figure}
    \vspace{-20mm}
    \hspace{-20mm}
    \includegraphics[width=200mm]{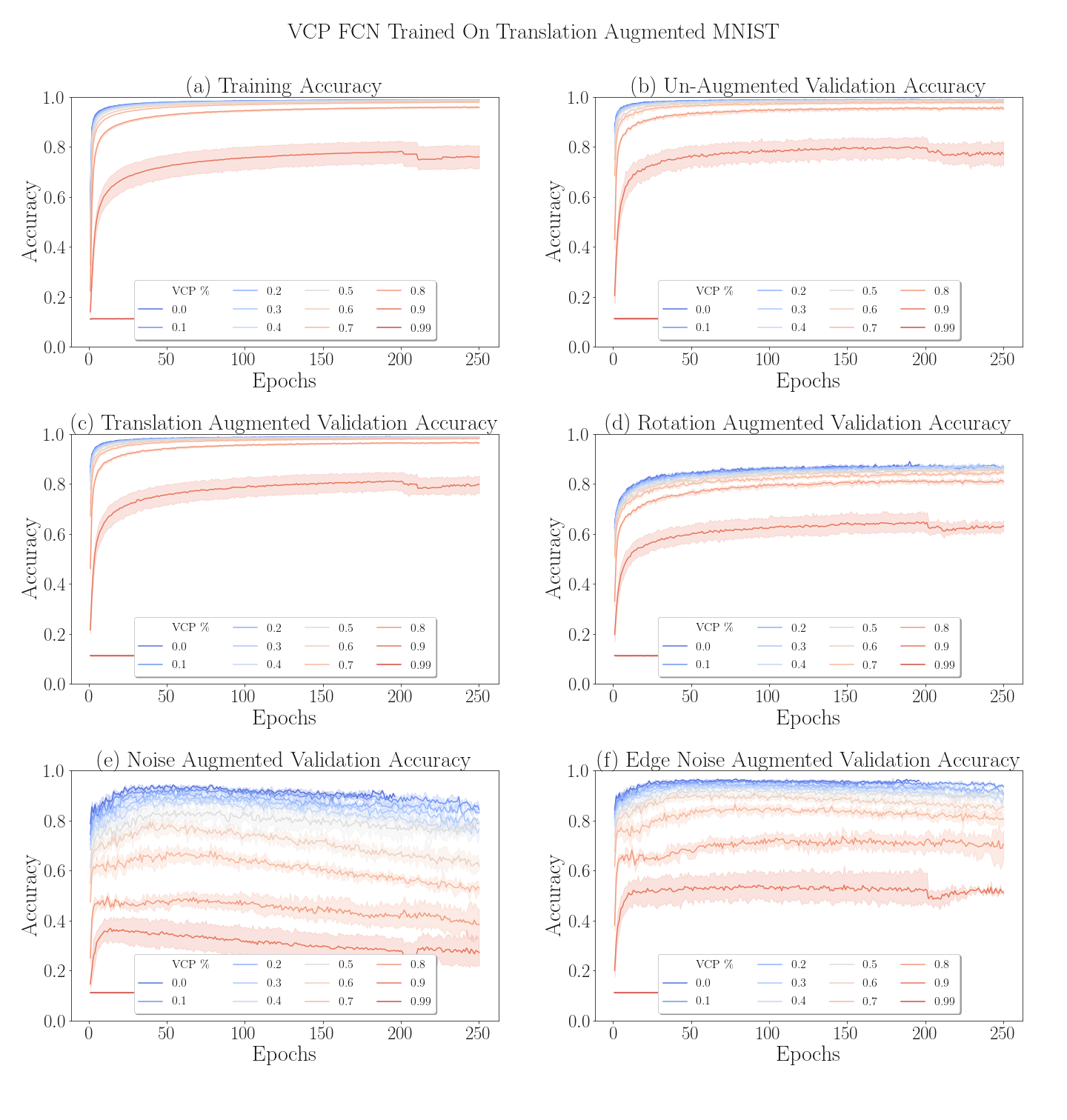}
    \caption*{C.1 FCNs trained with variable connection patterns on MNIST. The legends indicate the probability of absent dentritic connections in a FCN filter. (a) Accuracy on translation augmented training set, 30\% translations. (b) Accuracy on the un-augmented validation set. (c) Accuracy on the translation augmented validation set, using 25\% translations. (d) Accuracy on the rotation augmented validation set. (e) Accuracy on the noise augmented validation set. (f) Accuracy on the edge noise augmented validation set. }
\end{figure}

\begin{figure}
    \vspace{-20mm}
    \hspace{-20mm}
    \includegraphics[width=200mm]{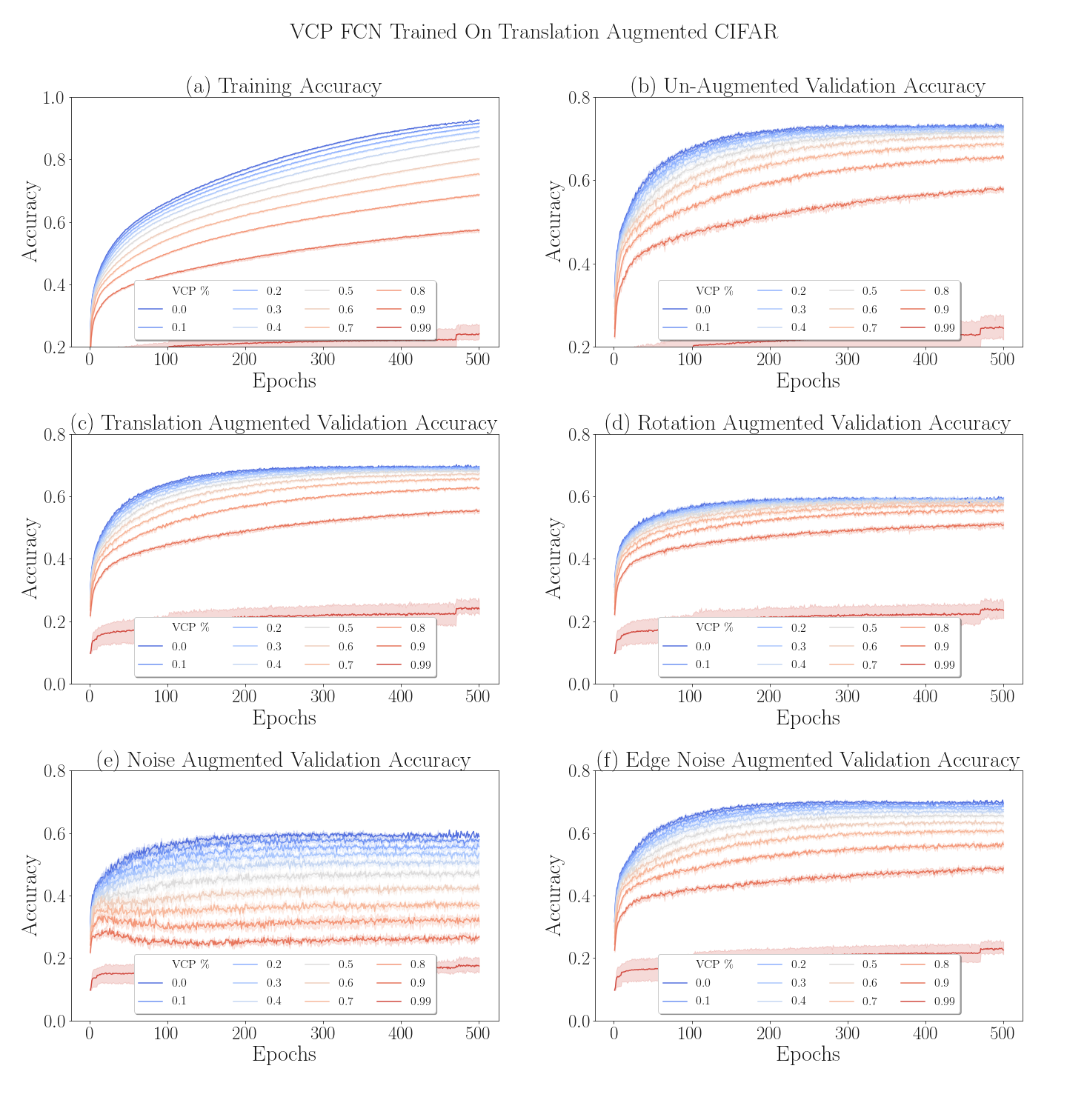}
    \caption*{C.2 FCNs trained with variable connection patterns on CIFAR. The legends indicate the probability of absent dentritic connections in a FCN filter. (a) Accuracy on translation augmented training set, 30\% translations. (b) Accuracy on the un-augmented validation set. (c) Accuracy on the translation augmented validation set, using 25\% translations. (d) Accuracy on the rotation augmented validation set. (e) Accuracy on the noise augmented validation set. (f) Accuracy on the edge noise augmented validation set.}
\end{figure}


\bibliographystyle{elsarticle-num}

\bibliography{sources,baldi,nn}

\end{document}